\documentclass[letterpaper]{article} % DO NOT CHANGE THIS
\usepackage{aaai25}  % DO NOT CHANGE THIS
\usepackage{times}  % DO NOT CHANGE THIS
\usepackage{helvet}  % DO NOT CHANGE THIS
\usepackage{courier}  % DO NOT CHANGE THIS
\usepackage[hyphens]{url}  % DO NOT CHANGE THIS
\usepackage{graphicx} % DO NOT CHANGE THIS
\def\UrlFont{\rm}  % DO NOT CHANGE THIS
\usepackage{natbib}  % DO NOT CHANGE THIS AND DO NOT ADD ANY OPTIONS TO IT
\usepackage{caption} % DO NOT CHANGE THIS AND DO NOT ADD ANY OPTIONS TO IT
\usepackage{algorithm}
\usepackage{algorithmic}

\usepackage{newfloat}
\usepackage{listings}
\DeclareCaptionStyle{ruled}{labelfont=normalfont,labelsep=colon,strut=off} % DO NOT CHANGE THIS
\floatstyle{ruled}
\newfloat{listing}{tb}{lst}{}
\floatname{listing}{Listing}
\usepackage{amsmath} % for math formatting
\usepackage{amssymb} % additional math symbols
\usepackage{booktabs} 

\usepackage{multirow}
\usepackage{graphicx}
\usepackage{enumitem}
\usepackage{svg}
\usepackage{amsmath}
\usepackage{graphicx}
\usepackage{booktabs}

\title{PALM: Pushing Adaptive Learning Rate Mechanisms for Continual Test-Time Adaptation}
\title{PALM: Pushing Adaptive Learning Rate Mechanisms for Continual Test-Time Adaptation}
\author {
    Sarthak Kumar Maharana,
    Baoming Zhang,
    Yunhui Guo
}
\affiliations {
    The University of Texas at Dallas, Richardson, USA\\
    \{sarthak.maharana, baoming.zhang,  yunhui.guo\}@utdallas.edu
}

\begin{document}

\maketitle

\begin{abstract}
Real-world vision models in dynamic environments face rapid shifts in domain distributions, leading to decreased recognition performance. Using unlabeled test data, continuous test-time adaptation (CTTA) directly adjusts a pre-trained source discriminative model to these changing domains. A highly effective CTTA method involves applying layer-wise adaptive learning rates for selectively adapting pre-trained layers. However, it suffers from the poor estimation of domain shift and the inaccuracies arising from the pseudo-labels. This work aims to overcome these limitations by identifying layers for adaptation via quantifying model prediction uncertainty without relying on pseudo-labels. We utilize the magnitude of gradients as a metric, calculated by backpropagating the KL divergence between the softmax output and a uniform distribution, to select layers for further adaptation. Subsequently, for the parameters exclusively belonging to these selected layers, with the remaining ones frozen, we evaluate their sensitivity to approximate the domain shift and adjust their learning rates accordingly. We conduct extensive image classification experiments on CIFAR-10C, CIFAR-100C, and ImageNet-C, demonstrating the superior efficacy of our method compared to prior approaches.
\end{abstract}

% \input{CameraReady/LaTeX/sec/introduction}
% \input{CameraReady/LaTeX/sec/related}
% \input{CameraReady/LaTeX/sec/proposed}
% \input{CameraReady/LaTeX/sec/results}
% \input{CameraReady/LaTeX/sec/conclusion}

% \input{CameraReady/LaTeX/supplementary}


\begin{links}
    \link{Code}{https://github.com/sarthaxxxxx/PALM}
    % \link{Datasets}{https://aaai.org/example/datasets}
    % \link{Extended version}{https://aaai.org/example/extended-version}
\end{links}

\section{Introduction}
\label{sec:intro}

In a real-world setting, machine-perception systems \cite{arnold2019survey} function in a rapidly changing environment. Pre-trained vision models are susceptible to performance degradation caused by potential distribution shifts in such contexts. For instance, while these models may have been trained on clean images, images captured by sensors could be corrupted by various weather conditions.

In the realm of addressing domain shifts due to image corruptions, \textit{test-time adaptation} (TTA) has been gaining a lot of traction lately \cite{dobler2023robust, wang2020tent}. Here, a pre-trained source model, with no access to its training (source) data due to privacy, is adapted to unlabeled test data through online training. One strategy in TTA involves mitigating domain shifts by adjusting the source model parameters using pseudo labels and minimizing the entropy of the model predictions \cite{wang2020tent}. While this approach effectively adapts the model to a single type of corruption, it faces two constraints when continually applied across a sequence of test tasks featuring different corruptions. Firstly, there is the issue of \textit{error accumulation} since pseudo labels are generated for unlabeled test data, errors accumulate over time. Secondly, \textit{catastrophic forgetting} occurs, resulting in a long-term loss of the model's source knowledge due to adaptations and large parameter updates across tasks.

\textit{Continual test-time adaptation} (CTTA) has been recently introduced to tackle these challenges. In \cite{niu2022efficient, gong2022robust, wang2020tent, schneider2020improving}, stable optimization is achieved by updating the batch normalization (BN) parameters or re-formulating its statistics. CoTTA \cite{wang2022continual} includes full model updates, with a teacher-student network setup, to handle the above issues. To further improve the performance of CoTTA, EcoTTA \cite{song2023ecotta} introduces the attachment of ``meta networks", \textit{i.e.}, a BN layer and a convolution block, to the frozen source model for efficient adaptation. 

More recently, the authors in \cite{lee2022surgical} demonstrated that based on the type of distributional shift of the test data, a certain subset of layers of a pre-trained model can be ``surgically" or heuristically chosen to fine-tune. Inspired by this finding, \textit{Layer-wise Auto-Weighting} (LAW) \cite{park2024layer} was introduced recently. LAW utilizes the Fisher Information Matrix (FIM) \cite{sagun2017empirical} to estimate domain shifts, automatically weighting trainable parameters and adjusting their learning rates (LRs) accordingly. While LAW achieves state-of-the-art results for CTTA, it still encounters two significant limitations: {\bf 1)} It presupposes that layer-wise importance can be accurately estimated via FIM with initially generated pseudo-labels. Nevertheless, it is widely recognized that pseudo-labels are inherently noisy and unreliable. {\bf 2)} Relying on directly accumulating such approximated importance from noisy labels to calculate domain shifts can lead to inaccuracies.

In this paper, we present a novel method called \textbf{PALM} to push the limits of adaptive learning rate methods for CTTA. To address catastrophic forgetting, our approach begins by identifying a subset of layers that significantly contribute to the uncertainty of model predictions. Rather than relying on pseudo-labels, we determine layer-wise importance by leveraging information from the gradient space, specifically by calculating the Kullback-Leibler (KL) divergence between the model output and a uniform distribution. This approach eliminates dependence on pseudo-labels and enables a more accurate estimation of layer-wise importance. Subsequently, the LRs of the parameters of these selected layers are adjusted based on the degree of domain shift observed in the test data. We use parameter sensitivity as an indicator of domain shift and employ weighted moving averages to aggregate sensitivity from previous test batches. Our approach provides a more accurate estimation of domain shift compared to existing adaptive LR methods, leading to more effective adjustments in LRs.

To summarise, our contributions are as follows: {\bf 1)} We present \textbf{PALM}, introducing novel improved schemes for CTTA with a specific focus on 
addressing the key limitations of existing adaptive learning rate methods. {\bf 2)} Considering the model's \textit{prediction uncertainty} at test-time, we automatically select and adapt layers requiring further updates while \textit{freezing} the rest. We adjust learning rates based on \textit{parameter sensitivity} to estimate domain shift, making the overall process \textit{computationally inexpensive}. {\bf 3)} Our method delivers superior results, via extensive evaluations, in both \textit{continual} and \textit{gradual TTA} across three standard benchmark datasets.

\section{Related Works}
\label{sec:related}

\noindent \textbf{Test-time adaptation (TTA)}: The goal of TTA is to adapt source models to unlabelled test data in real-time using a source-free online approach \cite{jain2011online, sun2020test, wang2020tent}. TENT \cite{wang2020tent} optimizes BN parameters by minimizing the Shannon entropy \cite{shannon1948mathematical} of pseudo labels. BN Stats Adapt (BN-1) \cite{schneider2020improving} replaces the BN statistics with those from corrupted test data, while AdaContrast \cite{chen2022contrastive} uses contrastive learning and a memory module to update the entire model. However, these methods address only single domain shifts and overlook continual shifts.

\noindent \textbf{Continual TTA (CTTA)}: CTTA requires models to adapt to dynamic domain shifts. CoTTA \cite{wang2022continual} uses a teacher-student setup with full model updates. The teacher model is updated using an exponential moving average (EMA) to deal with catastrophic forgetting and error accumulation with time. EcoTTA \cite{song2023ecotta} adds ``meta networks" to a frozen source model, computing an $L_{1}$ regularization loss to prevent catastrophic forgetting. DSS \cite{wang2024continual} introduces a dynamic thresholding framework to differentiate between low and high sample quality.

\noindent \textbf{Efficient model fine-tuning}: Freezing model parameters to learn information from downstream data has witnessed a good body of work \cite{sener2016learning, long2016unsupervised, zintgraf2019fast, guo2019spottune}. Depending upon the distribution shift of the target task, in \cite{lee2022surgical}, the authors demonstrate that heuristically selecting a subset of convolutional blocks for training while freezing the remaining blocks works better than full fine-tuning. 

\noindent  \textbf{Adaptive LR methods}: Research such as \cite{yosinski2014transferable, zeiler2014visualizing} demonstrates that deeper layers of a network extract task-specific features, unlike shallow layers. AutoLR \cite{ro2021autolr} developed an adaptive LR scheme based on weight variations between layers, although these rates tend to mostly stay constant during training. For CTTA, LAW \cite{park2024layer} uses the FIM \cite{sagun2017empirical} to assess domain shifts and adjust parameter LRs accordingly. However, using pseudo-labels to calculate FIM may lead to inaccuracies, and accumulating FIMs across domains can result in errors. Moreover, the adjustment of LR weights for each layer is constrained by the base LR, disregarding the relative ``sensitivity" of parameters to the adaptation needs of the current batch and domain.

\noindent  \textbf{Parameter sensitivity}: Neural network pruning literature extensively explores parameter sensitivity \cite{lecun1989optimal, luo2017thinet, theis2018faster, molchanov2019importance}. It suggests that less sensitive parameters require higher LRs for effective adaptation, and vice-versa. Additionally, \cite{liang2022no} introduces a concept of temporal sensitivity variation, influenced by uncertainty, as a factor in training models for tasks like natural language understanding and machine translation.

\section{Preliminary}
\label{sec:prelims}

\noindent \textbf{Problem setup}: CTTA aims to adapt a discriminative source model, parameterized as \(\theta^s\) and trained on the source dataset \((\mathcal{X}^{\text{source}}, \mathcal{Y}^{\text{source}})\), to multiple incoming tasks \(\mathcal{T}_i = \{x_k\}_{k=1}^K\), at test-time. Each \(\mathcal{T}_i\) signifies the \(i^{th}\) task with \(K\) batches, across varying data distributions. At time step \textit{t}=0, the model is initialized to ${\theta^s}$. It is then gradually adapted to each incoming batch $x_k$ of $\mathcal{T}_{i}$ in an \textbf{online} manner, where the model parameters are updated to ${\theta^t}$. The source dataset is unavailable in this adaptation process due to privacy and storage constraints. With the continuum of tasks, at each \(t\), the complete model parameters \(\theta^t\) consist of a deep tensor with \(N\) layers, expressed as \(\theta^t = \{\theta^t_1, \ldots, \theta^t_N\}\). Each layer \(\theta^t_n\) can represented by \(\theta^t_n = \{\theta^t_{1,n}, \ldots, \theta^t_{j,n}, \ldots \}\), where $\theta_{j,n}^t$ is an arbitrary parameter of layer $\theta_n^t$.

\noindent \textbf{Core motivations}: We provide a detailed view of \textbf{surgical fine-tuning (SFT)} \cite{lee2022surgical}, \textbf{AutoLR} \cite{ro2021autolr}, and \textbf{LAW} \cite{park2024layer}. It is critical to understand the design principles and reasons for having an adaptive LR scheme for CTTA, which \underline{motivates} this work.

To preserve the information during the source pre-training phase while fine-tuning on a target task of a different distribution like an image corruption, \textbf{SFT} empirically shows that \underline{heuristically} fine-tuning only the first convolutional block results in better performance over full fine-tuning. Even though it is a desirable outcome for efficient fine-tuning, in this work, we close the gap by \underline{automatically} selecting the layers that would adapt well to the target task.

Inspired by \cite{yosinski2014transferable, zeiler2014visualizing}, \textbf{AutoLR}, proposed for supervised model fine-tuning on a downstream task, claims that the weight changes across layers should be in an increasing order, where the weight change $\delta w_n$ in the $n^{th}$ layer with weights $w_n$ at training epoch \textit{e} is $\delta w_{n} = \| w_n^e\ - w_n^{(e-1)} \|$. They hypothesize that initial layers, which extract general features, should have minimal $\delta w_n$, while deeper layers extract task-specific features. An adaptive LR optimization scheme constrains weight variations to increase across layers, without assessing each layer's importance during adaptation. However, the LRs for each layer show minimal deviation and remain consistent throughout the training process.

As the first adaptive LR work for CTTA, \textbf{LAW} hypothesizes that identifying and controlling the adaptation rate of the different layers is useful \cite{lee2022surgical}. To understand the loss landscape and layer importance, the authors compute the gradient from the log-likelihood as $s(\theta_{j,n}^t; x_k)=\nabla_{\theta_{j,n}^t} \log (p(x_k))$, where $p(x_k)$ is the predicted output. Then, the FIM \cite{sagun2017empirical}, as a domain-shift indicator, is computed as a first-order approximation of the Hessian matrix. This layer-wise importance scheme completely relies on the pseudo-labels to compute the log-likelihood and is highly unreliable and uncertain. Due to the distribution shift, the label quality would begin to decrease for a continuum of tasks. The authors also propose to capture ``domain-level" FIM ($\hat{F_n^t}$), for each layer, by a direct accumulation of the current layer-wise FIM (${F_n^t}$) as, $\hat{F}_n^{t} = \hat{F}_n^{t-1} + F_n^{t}$. 

This means that LAW puts equal emphasis on $\hat{F}_n^{t}$ and ${F_n^t}$, for each layer. An issue arises when ${F}_n^{t}$ of the current batch is small but gets outweighed by a ``relatively" larger $\hat{F}_n^{t}$. This raises questions on the amount of adaptation required by the $n^{th}$ layer for batch $x_k$ as the FIMs dynamically change for each test batch. This leads to a possible imbalance between the accumulated FIMs $\hat{F}_n^{t}$ and the current FIM ${F_n^t}$. 

Drawing from these primary observations, our method focuses on addressing the following: \textbf{1)} Based on the prediction uncertainty of the model, we \underline{automatically} select a subset of layers that would require further adaptation to a task, without relying on the pseudo-labels. \textbf{2)} As a domain shift indicator, we leverage the idea of \underline{parameter sensitivity} of these layers only and adaptively alter their learning rates. We \underline{freeze} the parameters of the unselected layers.

\section{Proposed Methodology}
\label{prop}

\begin{figure}
\centering
\includegraphics[width=\columnwidth]{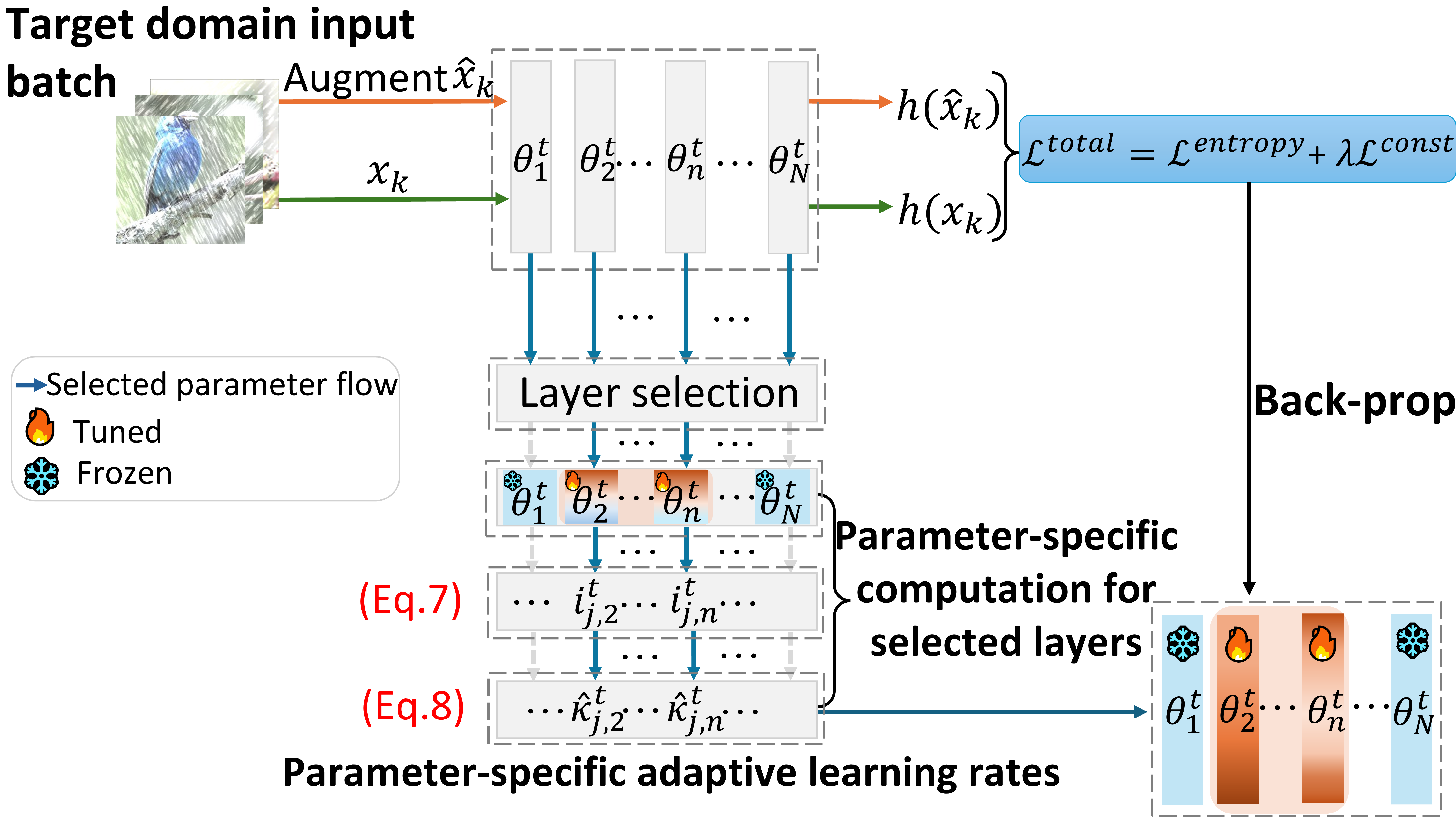}
\caption{The framework of our proposed method \textit{PALM}.}
\label{fig:aframework}
\end{figure}

Figure  \ref{fig:aframework} illustrates our approach in \textbf{two} stages.

\noindent \textbf{Layer selection by quantifying prediction uncertainty:}\label{pred_uncert}
CTTA maximizes model performance in the current domain and so it is vital to assess prediction certainty amidst rapid domain shifts. In the literature on continual learning and TTA, several works utilize FIM \cite{kirkpatrick2017overcoming, schwarz2018progress, park2024layer}. However, the unavailability of true labels, with unknown distribution, pushes us to capture the model prediction uncertainty and adapt selected layers, automatically chosen, to a batch of test data, unlike SFT \cite{lee2022surgical}.

Motivated by \cite{huang2021importance}, which proposed an uncertainty scoring function from the gradient space, we compute the gradients of each layer. In essence, we backpropagate the KL divergence between the softmax probabilities and a uniform distribution \textbf{u} = [$\frac{1}{C}$, \ldots, $\frac{1}{C}$] $\in$ $\mathbb{R}^{C}$, where $C$ is the total number of classes of the current task. This is a measurement of how far the distribution of the model's prediction is from a uniform distribution. Intuitively, the KL divergence will be small with a large distribution shift. Let $h(\cdot)$ denote the classifier of the model. The loss $\mathcal{L}(\theta^t)$, to compute the gradients, is computed as follows,
\begin{equation}\label{eq:initial}
    \mathcal{L}(\theta^t) = \text{KL}(\text{softmax}(h(x_k)/T) || \text{\textbf{u}})
\end{equation}
where the classifier's output logits $h(x_k)$ is smoothed by a temperature $T$. Let $\hat h(x_k)$ = $h(x_k)/T$. Our intuition is that since batches within the same task are of the same distribution, smoothing the logits with large values of $T$ will make them indistinguishable. With model parameters $\theta^t$, the gradients (${{\nabla_{\theta_{j,n}^t} \mathcal{L}({\theta}^t)}}$ = $\frac{\partial \mathcal{L}(\theta^t)}{\partial \theta_{j,n}^t}$) can be simplified to, 
\begin{equation}\label{eq:kl_final}
    {{\nabla_{\theta_{j,n}^t} \mathcal{L}({\theta}^t)}} = \frac{1}{C}\sum_{c=1}^{C}\frac{\partial \mathcal{L_\text{CE}}(\hat h(x_k),c)}{\partial \theta_{j,n}^t}
\end{equation}
Overall, the gradients boil down to the averaged cross-entropy loss between the smoothed output logits and a uniform label (Eq. \ref{eq:kl_final}). In CTTA, since ground-truth labels are absent, the gradients are more indicative of familiarity of the model with the batch of data. For batches within the same task, that share the same distribution, the vector norm of Eq. \ref{eq:kl_final} would be higher because of a higher KL divergence since the softmax predictions would be less uniformly distributed. However, if the norm is smaller, this would mean that the softmax outputs are closer to \textbf{u}. Our key idea is to select all the parameters of layers with the gradient norms to be less than or equal to a threshold $\eta$. \textit{Layers with small gradient norms would require more adaptation for the model to make a confident prediction. We freeze the unselected layers which also helps to alleviate catastrophic forgetting to a large extent by preserving the prior source knowledge.} So, the scoring function, for each layer's parameters, is defined as $\mathcal{Z}_{\theta_{j,n}^t} = \|{\nabla_{\theta_{j,n}^t} \mathcal{L}({\theta}^t)}\|_1$.
% \begin{equation}\label{eq:z_score}
%     \mathcal{Z}_{\theta_n} = \|{{\nabla \mathcal{L}({\theta}_n^t)}}\|_1
% \end{equation}
As mentioned earlier, $\mathcal{Z}_{\theta_{j,n}^t}$ $\le$ $\eta$ allows the model to adaptively determine which specific layers need further adaptation. We freeze the layers with $\mathcal{Z}_{\theta_{j,n}^t}$ $>$ $\eta$ \textit{i.e.}, set their LR to 0.

\noindent \textbf{Parameter sensitivity as a domain shift indicator}: New studies have found that model parameters demonstrate levels of criticality \cite{zhang2022all} across different positions. Moreover, not all parameters adapt uniformly, and removing some can lower test generalization errors \cite{mozer1988skeletonization, rasmussen2000occam}. Inspired by this observation, we assert that parameters contribute differently during CTTA. From the previous stage, we automatically select the layers based on the prediction uncertainty. \textit{For parameters of these selected layers only}, we compute their sensitivity \textit{i.e.}, an approximation of an error when it is removed. Motivated by \cite{molchanov2016pruning, molchanov2019importance}, we compute the sensitivity of a parameter $\theta_{j,n}^t$ based on the loss (Eq. \ref{eq:initial}) as follows, 
\begin{equation}\label{eqn:sens}
    S_{j,n}^t = \mathcal{L}({\theta}^t) - \mathcal{L}({\theta}^t | \theta_{j,n}^t = 0)
\end{equation}
Precisely, $S_{j,n}^t$ approximates the change in the loss landscape at $\theta^t$ when $\theta_{j,n}^t$ is removed. Sensitivity, computed as a ``weighting factor" of model weights \cite{molchanov2016pruning, molchanov2019importance, liang2022no}, determines the level of adaptation needed for model parameters. We utilize the gradients of parameters from the selected layers, computed by Eq. \ref{eq:kl_final}, to compactly derive Eq. \ref{eqn:sens} from the first-order Taylor series expansion of $\mathcal{L}(\theta^t)$ \cite{lee2018snip} to,
\begin{equation}\label{eqn:final_sens}
    S_{j,n}^t = |\theta_{j,n}^t \nabla_{\theta_{j,n}^t} \mathcal{L}({\theta^t})|
\end{equation}
Our rationale is as follows: parameters with low sensitivity are pushed to have higher LRs owing to more adaptation. However, $S_{j,n}^t$ of $\theta_{j,n}^t$ can be highly uncertain and would have a large variation. We introduce a method to capture this ``domain-level" uncertainty by computing a weighted moving average of the sensitivity $S_{j,n}^t$. This is formulated as,
\begin{equation}\label{eq:unc}
    \hat S_{j,n}^t = \alpha S_{j,n}^t + (1 - \alpha)\hat S_{j,n}^{(t-1)} 
\end{equation}
where $\hat S_{j,n}^t$ denotes the ``domain-level" sensitivity, computed on-the-fly, and $\alpha$ $\in$ [0,1] is a smoothing factor. Specifically, $\alpha$ decides the contribution of the current sensitivity $S_{j,n}^t$ to $\hat S_{j,n}^t$. Since CTTA happens at the batch level, it is critical to achieve a good balance between $S_{j,n}^t$ and $\hat S_{j,n}^t$, with a slightly larger focus on $S_{j,n}^t$. As can be seen, we do not directly accumulate the ``domain-level" information, unlike LAW. Due to large distribution shifts happening because of the continuum of tasks, larger deviations of $S_n^t$ from $\hat S_n^t$ are bound to happen. 
We quantify this statistic as an absolute difference,

\begin{equation}\label{eq:d_ind}
    \hat D_{j,n}^t = |S_{j,n}^t - \hat S_{j,n}^t|
\end{equation}
We emphasize that since Eq. \ref{eq:unc} leverages sensitivity from previously computed batches and domains, $\hat D_{j,n}^t$ would act as a good indicator of how uncertain $S_{j,n}^t$ is to the previous batches and domains \textit{i.e.} $\hat S_{j,n}^t$, for $\theta_{j,n}^t$. This would take care of error accumulation too. To automatically alter the LRs of these parameters, the ``importance" ($i_{j,n}^t$) is calculated as,
\begin{equation}\label{eqn:lr}
    i_{j,n}^t = \hat D_{j,n}^t / \hat S_{j,n}^t
\end{equation}
Eq. \ref{eq:d_ind} enforces $\hat D_{j,n}^t$$\ge$0 which, in turn, makes $i_{j,n}^t$$\ge$0. And so, the final learning rate to update $\theta_{j,n}^t$ would be,
\begin{equation}\label{eq:update_lr}
    \hat \kappa_{j,n}^t = \kappa * i_{j,n}^t
\end{equation}
where $\kappa$ is the base fixed LR. The intuition behind the formulation of Eq. \ref{eqn:lr} is - smaller values of $\hat S_{j,n}^t$ indicate lower sensitivity, and hence larger parameter updates are required via promoting the LR. At the same time, large uncertainties by $\hat D_{j,n}^t$ would also require larger updates. To avoid any overflow/underflow, we add $\epsilon$ (0$<$$\epsilon$$\ll$1) to the numerator and the denominator in Eq. \ref{eqn:lr}.

\noindent \textbf{Optimization}: Though our method is robust, it is necessary to avoid miscalibrated labels. TENT \cite{wang2020tent} minimized the entropy on the output logits, \textit{l}, following $\mathcal{H}(l) = -\sum_{c}^{}p(l_c)\log(p(l_c))$, where $p(l_c)$ is the probability for class \textit{c}. Following EcoTTA \cite{song2023ecotta}, we use entropy minimization as the objective for adaptation where $\mathcal{H}(l)$ is minimized for samples below a certain fixed threshold $\mathcal{H}_0$ as,
\begin{equation}
    \mathcal{L}^{entropy} = \mathbf{1}_{\mathcal{H}(l) \le \mathcal{H}_0}\mathcal{H}(l)
\end{equation}
$\mathcal{H}_0$ is set to $0.4 \times \log C$ following EcoTTA. For a fair comparison with prior works \cite{wang2022continual, park2024layer}, we apply a consistency loss between \textit{l} and $\hat{l}$, computed as the output logits from the augmented version of $x_k$, to regularize the model as follows,
\begin{equation}
    \mathcal{L}^{const} = -\sum_{c}^{}p(l_c)\log(p(\hat l_c))
\end{equation}
where, $p(\hat l_c)$ is the probability for class \textit{c}. Our overall optimization objective is as follows, 
\begin{equation}\label{eq:final_optim}
    \mathcal{L}^{total} = \mathcal{L}^{entropy} + \lambda\mathcal{L}^{const}
\end{equation}
where $\lambda$ is a regularisation coefficient. 
In summary, upon calculating the parameter-specific LRs (Eq. \ref{eq:update_lr}), we optimize the continually fine-tuned model to minimize Eq. \ref{eq:final_optim}.

\section{Experiments and Results}
\label{setup}

\begin{table*}[!t]
\centering
\footnotesize
\begin{tabular}{
@{}l
@{\hspace{0.8pt}}c
@{\hspace{0.8pt}}c
@{\hspace{0.8pt}}c
@{\hspace{0.8pt}}c
@{\hspace{0.8pt}}c
@{\hspace{0.8pt}}c
@{\hspace{0.8pt}}c
@{\hspace{0.8pt}}c
@{\hspace{0.8pt}}c
@{\hspace{0.8pt}}c
@{\hspace{0.8pt}}c
@{\hspace{0.8pt}}c
@{\hspace{0.8pt}}c
@{\hspace{0.8pt}}c
@{\hspace{0.8pt}}c
@{\hspace{0.8pt}}c@{}}
\toprule
\textbf{Method} & \textbf{Gaussian} & \textbf{Shot} & \textbf{Impulse} & \textbf{Defocus} & \textbf{Glass} & \textbf{Motion} & \textbf{Zoom} & \textbf{Snow} & \textbf{Frost} & \textbf{Fog} & \textbf{Brightness} & \textbf{Contrast} & \textbf{\begin{tabular}[c]{@{}c@{}}Elastic\\Transform\end{tabular}} & \textbf{Pixelate} & \textbf{JPEG} & \textbf{Mean} \\
\midrule
Source         & 72.3 & 65.7 & 72.9 & 46.9 & 54.3 & 34.8 & 42.0 & 25.1 & 41.3 & 26.0 & 9.3 & 46.7 & 26.6 & 58.5 & 30.3 & 43.5 \\
BN-1           & 28.1 & 26.1 & 36.3 & 12.8 & 35.3 & 14.2 & 12.1 & 17.3 & 17.4 & 15.3 & 8.4 & 12.6 & 23.8 & 19.7 & 27.3 & 20.4 \\
Tent Cont. & 24.8 & 20.6 & 28.6 & 14.4 & 31.1 & 16.5 & 14.1 & 19.1 & 18.6 & 18.6 & 12.2 & 20.3 & 25.7 & 20.8 & 24.9 & 20.7 \\
CoTTA          & 24.3 & 21.3 & 26.6 & 11.6 & 27.6 & 12.2 & 10.3 & 14.8 & 14.1 & 12.4 & 7.5 & 10.6 & 18.3 & 13.4 & 17.3 & 16.2 \\
EcoTTA         & 23.8 & 18.7 & 25.7 & 11.5 & 29.8 & 13.3 & 11.3 & 15.3 & 15.0 & 13.0 & 7.9 & 11.3 & 20.2 & 15.1 & 20.5 & 16.8 \\
AdaContrast    & 29.1 & 22.5 & 30.0 & 14.0 & 32.7 & 14.1 & 12.0 & 16.6 & 14.9 & 14.4 & 8.1 & 10.0 & 21.9 & 17.7 & 20.0 & 18.5 \\
SFT (cont.)& 27.7 & 24.1 & 33.8 & 12.2 & 33.1 & 13.1 & 10.9 & 15.9 & 15.6 & 13.8 & 7.8 & 10.8 & 21.2 & 16.6 & 22.9 & 18.6 \\
DSS            & 24.1 & 21.3 & 25.4 & 11.7 & 26.9 & 12.2 & 10.5 & 14.5 & 14.1 & 12.5 & 7.8 & 10.8 & 18.0 & 13.1 & 17.3 & 16.0 \\
LAW            & 24.7 & 18.9 & 25.5 & 12.9 & 26.7 & 15.0 & 11.8 & 15.1 & 14.7 & 15.9 & 10.1 & 13.8 & 19.4 & 14.7 & 18.3 & 17.2 \\
Ours           & 25.8 & 18.1 & 22.7 & 12.3 & 25.3 & 13.1 & 10.7 & 13.5 & 13.1 & 12.2 & 8.5 & 11.8 & 17.9 & 12.0 & 15.4 & \textbf{15.5} \\
\bottomrule
\end{tabular}
\caption{\textit{Mean errors (\%) on CIFAR-10C} - CTTA mean errors of the 15 corruptions (tasks) at a severity level of 5.}
\label{table:cifar10ctta}
\end{table*}

\begin{table*}[!t]
\centering
\footnotesize
\begin{tabular}{
@{}l
@{\hspace{0.8pt}}c
@{\hspace{0.8pt}}c
@{\hspace{0.8pt}}c
@{\hspace{0.8pt}}c
@{\hspace{0.8pt}}c
@{\hspace{0.8pt}}c
@{\hspace{0.8pt}}c
@{\hspace{0.8pt}}c
@{\hspace{0.8pt}}c
@{\hspace{0.8pt}}c
@{\hspace{0.8pt}}c
@{\hspace{0.8pt}}c
@{\hspace{0.8pt}}c
@{\hspace{0.8pt}}c
@{\hspace{0.8pt}}c
@{\hspace{0.8pt}}c@{}}
\toprule
\textbf{Method} & \textbf{Gaussian} & \textbf{Shot} & \textbf{Impulse} & \textbf{Defocus} & \textbf{Glass} & \textbf{Motion} & \textbf{Zoom} & \textbf{Snow} & \textbf{Frost} & \textbf{Fog} & \textbf{Brightness} & \textbf{Contrast} & \textbf{\begin{tabular}[c]{@{}c@{}}Elastic\\Transform\end{tabular}} & \textbf{Pixelate} & \textbf{JPEG} & \textbf{Mean} \\
\midrule
Source         & 73.0 & 68   & 39.4 & 29.3 & 54.1 & 30.8 & 28.8 & 39.5 & 45.8 & 50.3 & 29.5 & 55.1 & 37.2 & 74.7 & 41.2 & 46.4 \\
BN-1           & 42.1 & 40.7 & 42.7 & 27.6 & 41.9 & 29.7 & 27.9 & 34.9 & 35.0 & 41.5 & 26.5 & 30.3 & 35.7 & 32.9 & 41.2 & 35.4 \\
Tent Cont. & 37.2 & 35.8 & 41.7 & 37.9 & 51.2 & 48.3 & 48.5 & 58.4 & 63.7 & 71.1 & 70.4 & 82.3 & 88.0 & 88.5 & 90.4 & 60.9 \\
CoTTA          & 40.1 & 37.7 & 39.7 & 26.9 & 38.0 & 27.9 & 26.4 & 32.8 & 31.8 & 40.3 & 24.7 & 26.9 & 32.5 & 28.3 & 33.5 & 32.5 \\
AdaContrast    & 42.3 & 36.8 & 38.6 & 27.7 & 40.1 & 29.1 & 27.5 & 32.9 & 30.7 & 38.2 & 25.9 & 28.3 & 33.9 & 33.3 & 36.2 & 33.4 \\
SFT (cont.)& 40.7 & 36.3 & 37.8 & 25.5 & 39.0 & 27.6 & 25.2 & 32.2 & 30.4 & 37.3 & 23.7 & 26.8 & 32.3 & 27.6 & 37.2 & 32.0 \\
DSS            & 39.7 & 36.0 & 37.2 & 26.3 & 35.6 & 27.5 & 25.1 & 31.4 & 30.0 & 37.8 & 24.2 & 26.0 & 30.0 & 26.3 & 31.1 & 30.9 \\
LAW            & 41.0 & 36.7 & 38.3 & 25.6 & 37.0 & 27.8 & 25.2 & 30.7 & 30.0 & 37.3 & 24.4 & 27.6 & 31.2 & 27.8 & 34.9 & 31.7 \\
Ours           & 37.3 & 32.5 & 34.9 & 26.2 & 35.3 & 27.5 & 24.6 & 28.8 & 29.1 & 34.1 & 23.5 & 26.9 & 31.2 & 26.6 & 34.1 & \textbf{30.1} \\
\bottomrule
\end{tabular}
\caption{\textit{Mean errors (\%) on CIFAR-100C} - CTTA mean errors of the 15 corruptions (tasks) at a severity level of 5.}
\label{table:cifar100ctta}
\end{table*}

% We contrast our proposed method against other state-of-the-art works in a CTTA setting. We also report results in a gradual test-time adaptation (GTTA) setting to broaden our study, following \cite{wang2022continual, park2024layer}. 
We evaluate our proposed method in a CTTA setting, comparing it to state-of-the-art approaches, and extend our study to the gradual test-time adaptation (GTTA) setting following \cite{wang2022continual, park2024layer}.

\subsection{Experimental Settings and Baselines}
\noindent \textbf{Datasets and Tasks}: Following the standard benchmarks set by \cite{wang2022continual}, we evaluate our proposed method on CIFAR-10C, CIFAR-100C, and ImageNet-C, based on image corruption schemes as set in \cite{hendrycks2019benchmarking}. Each dataset contains a set of 15 corruption styles as tasks (\textit{e.g.} gaussian noise, shot noise, $\ldots$) with 5 severity levels. We follow CoTTA and apply the same test-time augmentations, like color jitter, gaussian blur, gaussian noise, random flips, and random affine, to the input $x_k$. As the evaluation metric, we report the mean classification error.

% in line with established practices outlined in \cite{wang2022continual, dobler2023robust}.

% \noindent \textbf{CTTA implementation details}: We evaluate our approach following prior SOTA methods \cite{wang2020tent, wang2022continual, chen2022contrastive, song2023ecotta, schneider2020improving, park2024layer}. 
\noindent \textbf{CTTA implementation details}: To maintain fairness regarding the source models, we employ WideResNet-28 (36.4M params) \cite{zagoruyko2016wide} for CIFAR-10C, ResNeXt-29 (6.8M params) \cite{xie2017aggregated} for CIFAR-100C, and ResNet-50 (23.5M params) \cite{he2016deep} for ImageNet-C, all available on RobustBench \cite{croce2020robustbench}. We optimize using an Adam optimizer, setting base learning rates ($\kappa$) to 5e-4 for CIFAR-10C and CIFAR-100C, and 5e-5 for ImageNet-C. For balancing parameter sensitivities, we set $\alpha$ to 0.5, 0.9, and 0.5 respectively, with temperature coefficients \textit{T} set to 50, 100, and 1000 respectively. We set $\eta$ to 1, 0.5, and 0.3, and $\lambda$ to 0.01 throughout. Batch sizes are set to 200, 200, and 64 for each dataset, following CoTTA, and results are only reported for the highest severity level of 5 for each task.

\noindent \textbf{GTTA implementation details}: \label{gtta_desc} In the standard CTTA setup, each task only receives images with the highest severity. However, in GTTA, following CoTTA, for each task, the severity levels are gradually changed as $\ldots$2$\rightarrow$1 $\xrightarrow[change]{task}$ 1$\rightarrow$2$\rightarrow$3$\rightarrow$4$\rightarrow$5$\rightarrow$4$\rightarrow$3$\rightarrow$2$\rightarrow$1 $\xrightarrow[change]{task}$ 1$\rightarrow$2$\ldots$. We adhere to the details specified in the CTTA setting.

\noindent \textbf{Baselines}: We compare against several CTTA baselines, including BN Stats Adapt (BN-1) \cite{schneider2020improving}, Tent Cont. \cite{wang2020tent}, CoTTA \cite{wang2022continual}, EcoTTA \cite{song2023ecotta}, AdaContrast \cite{chen2022contrastive}, DSS \cite{wang2024continual}, and LAW \cite{park2024layer}. We also contrast our work against surgical fine-tuning \cite{lee2022surgical}. For EcoTTA, we report the results on CIFAR-10C and ImageNet-C only since they do not use ResNeXt-29 \cite{xie2017aggregated} for CIFAR-100C. Following the findings of \cite{lee2022surgical}, we continually fine-tune only the first convolutional block of the respective models, denoted as ``SFT (cont.)", using an SGD optimizer with fixed learning rates of 1e-3 for the CIFAR datasets and 5e-4 for ImageNet-C. For the results of LAW, we reproduce their results strictly following their official implementation \footnote{https://github.com/junia3/LayerwiseTTA/tree/main} with parameters as mentioned in their paper.

\begin{table*}[!t]
\centering
\footnotesize
\begin{tabular}{
@{}l
@{\hspace{0.8pt}}c
@{\hspace{0.8pt}}c
@{\hspace{0.8pt}}c
@{\hspace{0.8pt}}c
@{\hspace{0.8pt}}c
@{\hspace{0.8pt}}c
@{\hspace{0.8pt}}c
@{\hspace{0.8pt}}c
@{\hspace{0.8pt}}c
@{\hspace{0.8pt}}c
@{\hspace{0.8pt}}c
@{\hspace{0.8pt}}c
@{\hspace{0.8pt}}c
@{\hspace{0.8pt}}c
@{\hspace{0.8pt}}c
@{\hspace{0.8pt}}c@{}}
\toprule
\textbf{Method} & \textbf{Gaussian} & \textbf{Shot} & \textbf{Impulse} & \textbf{Defocus} & \textbf{Glass} & \textbf{Motion} & \textbf{Zoom} & \textbf{Snow} & \textbf{Frost} & \textbf{Fog} & \textbf{Brightness} & \textbf{Contrast} & \textbf{\begin{tabular}[c]{@{}c@{}}Elastic\\Transform\end{tabular}} & \textbf{Pixelate} & \textbf{JPEG} & \textbf{Mean} \\
\midrule
Source         & 97.8 & 97.1 & 98.2 & 81.7 & 89.8 & 85.2 & 78.0 & 83.5 & 77.1 & 75.9 & 41.3 & 94.5 & 82.5 & 79.3 & 68.6 & 82.0 \\
BN-1           & 85.0 & 83.7 & 85.0 & 84.7 & 84.3 & 73.7 & 61.2 & 66.0 & 68.2 & 52.1 & 34.9 & 82.7 & 55.9 & 51.3 & 59.8 & 68.6 \\
Tent Cont. & 81.6 & 74.6 & 72.7 & 77.6 & 73.8 & 65.5 & 55.3 & 61.6 & 63.0 & 51.7 & 38.2 & 72.1 & 50.8 & 47.4 & 53.3 & 62.6 \\
CoTTA          & 84.7 & 82.1 & 80.6 & 81.3 & 79.0 & 68.6 & 57.5 & 60.3 & 60.5 & 48.3 & 36.6 & 66.1 & 47.2 & 41.2 & 46.0 & 62.7 \\
EcoTTA         & - & - & - & - & - & - & - & - & - & - & - & - & - & - & - & 63.4 \\
AdaContrast    & 82.9 & 80.9 & 78.4 & 81.4 & 78.7 & 72.9 & 64.0 & 63.5 & 64.5 & 53.5 & 38.4 & 66.7 & 54.6 & 49.4 & 53.0 & 65.5 \\
SFT (cont.)& 83.4 & 79.5 & 77.2 & 86.1 & 82.4 & 72.2 & 61.1 & 63.1 & 64.7 & 51.2 & 35.9 & 72.9 & 56.2 & 46.7 & 50.2 & 65.5 \\
DSS            & 84.6 & 80.4 & 78.7 & 83.9 & 79.8 & 74.9 & 62.9 & 62.8 & 62.9 & 49.7 & 37.4 & 71.0 & 49.5 & 42.9 & 48.2 & 64.6 \\
LAW            & 81.8 & 75.0 & 72.1 & 77.4 & 73.2 & 63.9 & 54.6 & 57.9 & 61.0 & 50.1 & 36.3 & 68.6 & 49.0 & 46.2 & 49.1 & 61.0 \\
Ours           & 81.1 & 73.3 & 70.8 & 77.0 & 71.8 & 62.3 & 53.9 & 56.7 & 60.8 & 50.4 & 36.2 & 65.9 & 48.0 & 45.2 & 48.0 & \textbf{60.1} \\
\bottomrule
\end{tabular}
\caption{\textit{Mean errors (\%) on ImageNet-C} - CTTA mean errors of the 15 corruptions (tasks) at a severity level of 5.}
\label{table:imagenetctta}
\end{table*}

\subsection{CTTA Classification Results}
{\bfseries Results on CIFAR-10C and CIFAR-100C:} We report the CTTA results in Tables \ref{table:cifar10ctta} and \ref{table:cifar100ctta}. Our method performs the best across all training setups and baselines for both datasets. We achieve large improvements of 4.9\% and 5.2\% for CIFAR-10C and 5.3\% and 30.8\% for CIFAR-100C, over BN-1 and Tent Cont. respectively. 
Concerning recent works of LAW and DSS, our method improves by 1.7\% and 0.5\% on CIFAR-10C and 1.6\% and 0.8\% on CIFAR-100C, respectively. Tent Cont. excels initially but accumulates errors over time, increasing mean error. In CoTTA, the method's reliance on full model updates renders it computationally expensive and unsuitable for deployment. An interesting observation is that with a drastic reduction in the number of trainable parameters, as in SFT (cont.), the mean errors are better than most CTTA benchmarks, if not the best. LAW uses the uncertain pseudo labels to calculate the domain shift approximated via the FIM. The minor error that is introduced for each batch begins to propagate across all the domains. In addition, the imbalance caused by the ``domain-level" FIM and the current batch's FIM, results in a higher error rate for the initial tasks. In our method, the computation of the model prediction uncertainty to automatically select layers is useful (Eq. \ref{eq:initial}). This confirms the findings in SFT \cite{lee2022surgical} that concentrated adaptation, for image corruptions, in the initial layers suffices.

{\noindent \bfseries Results on Imagenet-C:} In Table \ref{table:imagenetctta}, we report the CTTA results on the ImageNet-C dataset. As shown, our method outperforms all the previous CTTA baselines and SFT (cont.). We observe an improvement of 0.9\% and 4.5\% over LAW and DSS respectively. In addition, we achieve 2.6\%, 5.4\%, and 2.5\% improvements over CoTTA, AdaContrast, and Tent Cont., respectively.

\begin{table}[htb!]
\scriptsize
\centering
\begin{tabular}{
@{}l
@{\hspace{2pt}}c
@{\hspace{2pt}}c
@{\hspace{2pt}}c
@{\hspace{2pt}}c
@{\hspace{2pt}}c
@{\hspace{2pt}}c
@{\hspace{2pt}}c
@{\hspace{2pt}}c@{}}
\toprule
\, \, \textbf{Dataset} & \textbf{Source} & \textbf{\begin{tabular}[c]{@{}c@{}}BN-1\end{tabular}} & \textbf{\begin{tabular}[c]{@{}c@{}}TENT\\Cont.\end{tabular}} & \textbf{CoTTA} & \textbf{AdaContrast} & \textbf{\begin{tabular}[c]{@{}c@{}}SFT\\(cont.)\end{tabular}} & \textbf{LAW} & \textbf{Ours} \\
\midrule
CIFAR-10C  & 24.7 & 13.7 & 20.4 & 10.9 & 12.1 & 11.3 & 10.9 & \textbf{9.5} \\
CIFAR-100C & 33.6 & 29.9 & 74.8 & 26.3 & 33.0 & \textbf{25.3} & 27.2 & 26.1 \\
ImageNet-C & 58.4 & 48.3 & 46.4 & 38.8 & 66.3 & 41.4 & 40.5 & \textbf{38.5} \\
\bottomrule
\end{tabular}
\caption{\textit{Mean errors (in \%) on CIFAR-10C, CIFAR-100C, and ImageNet-C} - GTTA mean errors averaged across all severity levels (1$\rightarrow$2$\rightarrow$3$\rightarrow$4$\rightarrow$5).}
\label{tab:gtta}
\end{table}

\subsection{GTTA Classification Results}
{\bfseries Results:} Here, we present the GTTA results on the benchmark datasets. For each task, the corruption severity of the inputs is gradually changed as discussed earlier. Table \ref{tab:gtta} reports the mean classification error for GTTA across all severities. We observe that, for all the datasets, our approach outperforms or performs comparably well across all the reported baselines. With most parameters frozen during the adaptation on each batch, our method is more robust and inexpensive than CoTTA, which also demonstrates competitive improvements in the mean classification error.

\subsection{Ablation Studies and Analysis}

\noindent {\bfseries Computational efficiency and Catastrophic forgetting (CF):} In our PALM, we first select a subset of layers from the gradient space (Eq. \ref{eq:kl_final}) and adapt them while strictly freezing the unselected layers. In Table \ref{tab:comp_eff}, we report the average \% of adapted model parameters per task with respect to the source model, for our method and CoTTA, across the benchmark datasets. Our method is computationally inexpensive and on an average updates 4.34\%, 6.65\%, and 17.21\% of their respective source parameters. This showcases significant computational advantages over CoTTA, which updates 100\% of the parameters. With a single NVIDIA A5000 GPU, PALM incurs a slightly higher adaptation time/batch, due to the two-stage proposed method. For example, on CIFAR-10, PALM runs for 0.8s per batch, whereas SFT (cont.) and LAW take 0.45 s and 0.73 s respectively. Our method alleviates CF by only adapting a small subset of model parameters, ensuring that much of the prior source knowledge is preserved. Quantitatively, the source models achieved zero-shot mean accuracies of 94.03\%, 76.06\%, and 75.19\% on respective datasets. After adaptation, PALM retained source knowledge better, with mean accuracies of 92.29\%, 77.46\%, and 70.87\%, compared to LAW's 91.67\%, 77.69\%, and 71.57\%. Despite the accuracy decline with more tasks, especially after adaptation on ImageNet-C, PALM matches or outperforms LAW.

\begin{table}[!htb]
% \centering
\scriptsize
\begin{minipage}[t]{0.4\linewidth}
\centering
\begin{tabular}{
@{}l
@{\hspace{1pt}}c
@{\hspace{1pt}}c@{}}
\toprule
\, \, \textbf{Dataset} & \textbf{Ours} & \textbf{CoTTA} \\
\midrule
CIFAR-10C  & \textbf{4.34} & 100  \\
CIFAR-100C & \textbf{6.65}  & 100   \\
ImageNet-C & \textbf{17.21} & 100  \\
\bottomrule
\end{tabular}
\caption{The average \% of adapted source parameters/task.}
\label{tab:comp_eff}
\end{minipage}%
% \hfill
\, \, 
\begin{minipage}[t]{0.6\linewidth}
% \centering
\begin{tabular}{
@{}l
@{\hspace{1pt}}c
@{\hspace{1pt}}c
@{\hspace{1pt}}c
@{\hspace{1pt}}c@{}}
\toprule
\, \, \textbf{Dataset} & \textbf{Ours} & \textbf{LARS} & \textbf{Prodigy} & \textbf{Adafactor} \\
\midrule
CIFAR-10C  & \textbf{15.5} & 18.9 & 20.4 & 73.8 \\
CIFAR-100C & \textbf{30.1} & 34.1 & 35.5 & 97.7 \\
ImageNet-C & \textbf{60.1} & 62.0 & 68.6 & 99.2 \\
\bottomrule
\end{tabular}
\caption{CTTA mean errors (\%) using adaptive LR optimizers vs. ours.}
\label{tab:optim}
\end{minipage}
\end{table}

% \begin{table}[!t]
% \centering
% \scriptsize
% \centering
% \begin{tabular}{
% @{}l
% @{\hspace{2pt}}c
% @{\hspace{2pt}}c
% @{\hspace{2pt}}c
% @{\hspace{2pt}}c@{}}
% \toprule
% \, \, \textbf{Dataset} & \textbf{Ours} & \textbf{LARS} & \textbf{Prodigy} & \textbf{Adafactor} \\
% \midrule
% CIFAR-10C  & \textbf{15.5} & 18.9 & 20.4 & 73.8 \\
% CIFAR-100C & \textbf{30.1} & 34.1 & 35.5 & 97.7 \\
% ImageNet-C & \textbf{60.1} & 62.0 & 68.6 & 99.2 \\
% \bottomrule
% \end{tabular}
% \caption{CTTA mean errors (\%) using adaptive LR optimizers vs. ours.}
% \label{tab:optim}
% \end{table}

\noindent {\bfseries Adaptive LR optimizers vs Ours:} Here, we study the effect of directly using adaptive LR optimizers like LARS \cite{you2017large}, Prodigy \cite{mishchenko2023prodigy}, and Adafactor \cite{shazeer2018adafactor}, for CTTA. We report the results in Table \ref{tab:optim}. The mentioned optimizers do not adjust layer-wise LRs based on parameter sensitivity, failing to capture rapid domain shifts and resulting in poorer performance compared to our method.

\begin{figure}[htb!]
% \centering
\begin{tabular}{
@{}c
@{\hspace{0.1pt}}c
@{\hspace{0.1pt}}c@{}}
{\footnotesize \,\,\,\,\, \, \, \, \textbf{WideResNet-28}} 
& {\footnotesize \,\,\,\,\,\, \, \, \, \textbf{ResNeXt-29}}
& {\footnotesize \,\,\,\,\,\, \, \, \, \textbf{ResNet-50}} \\
\includegraphics[width=0.33\columnwidth]{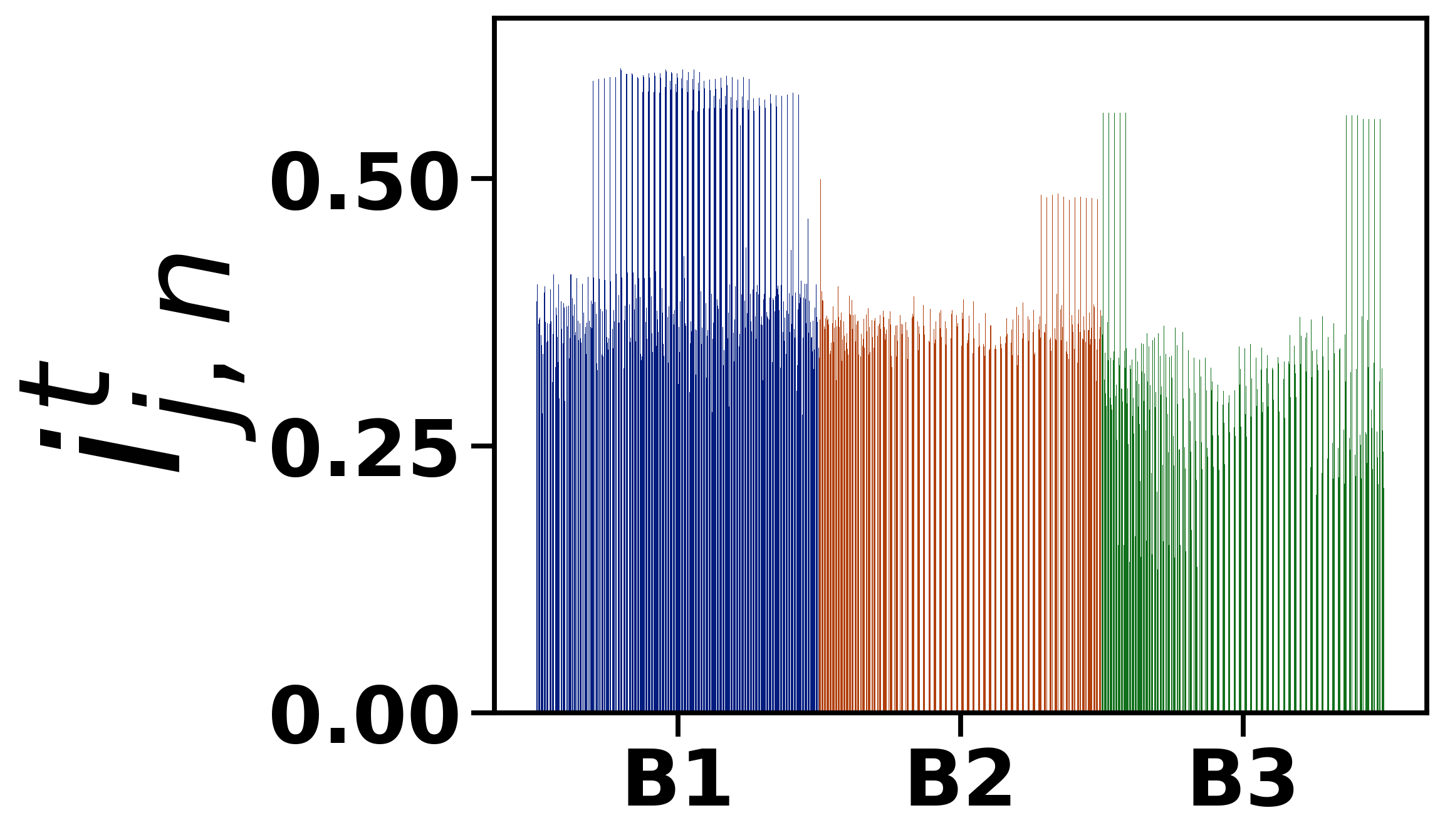}
& \includegraphics[width=0.33\columnwidth]{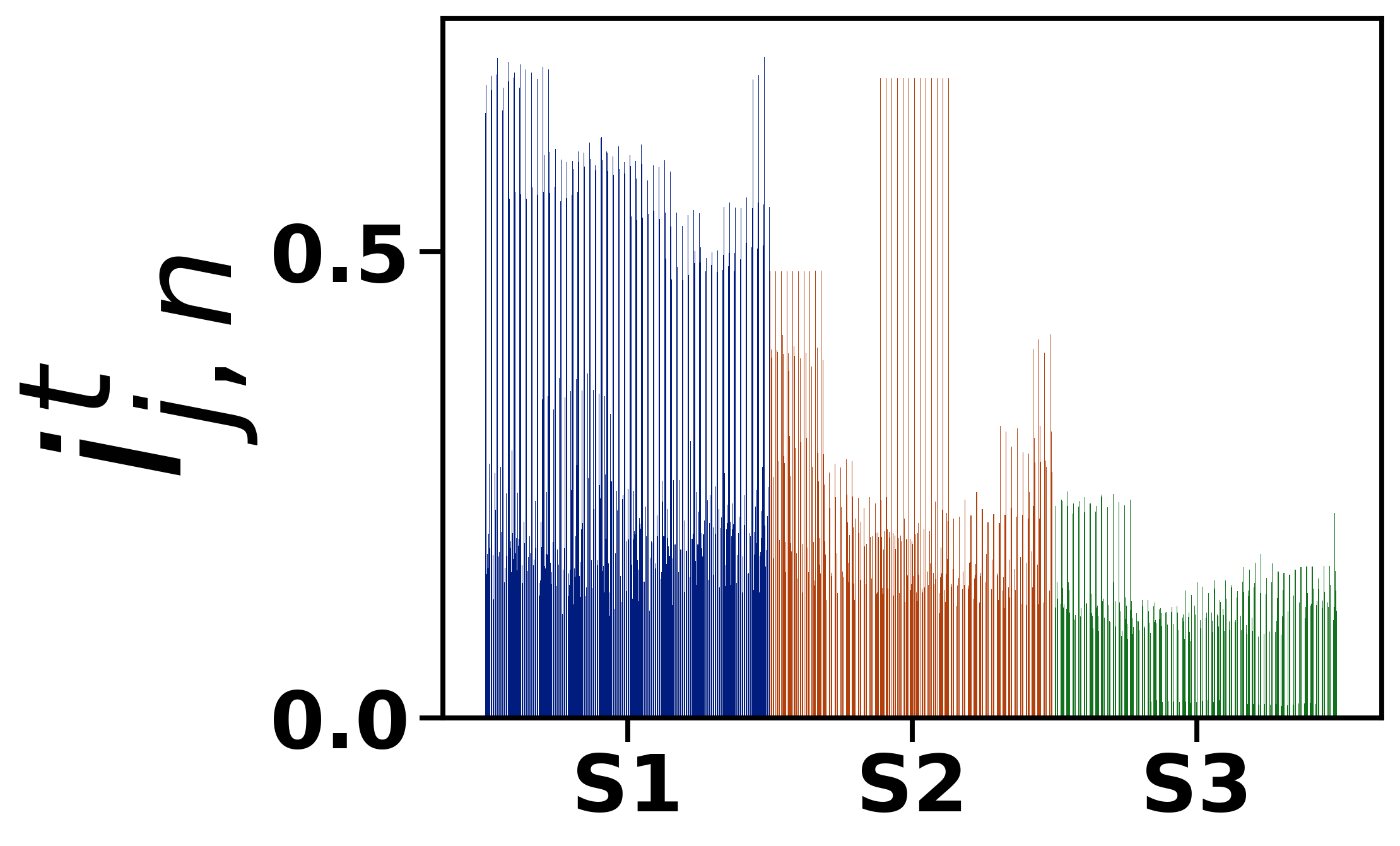}
& \includegraphics[width=0.33\columnwidth]{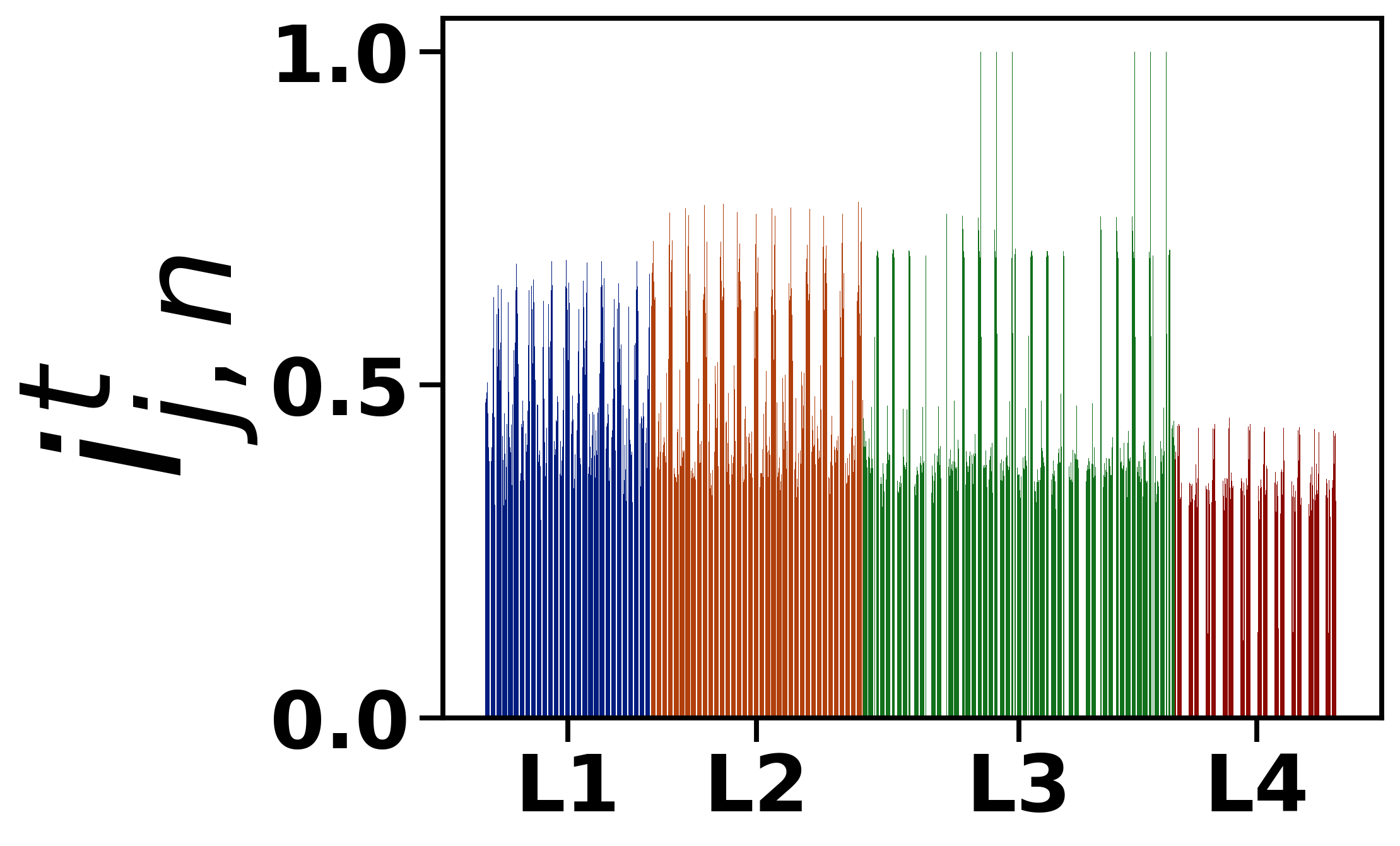} \\
\includegraphics[width=0.33\columnwidth]{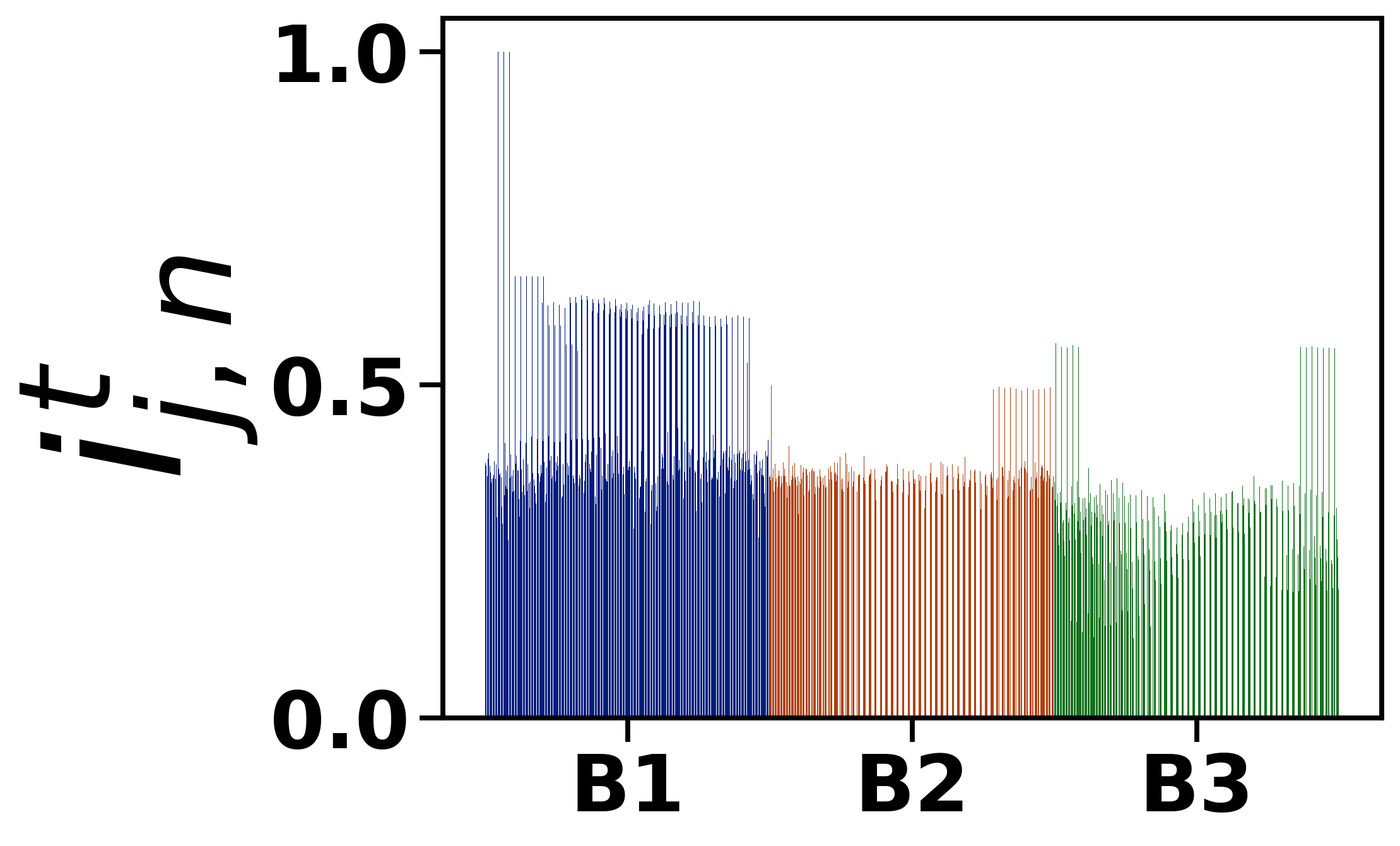}
& \includegraphics[width=0.33\columnwidth]{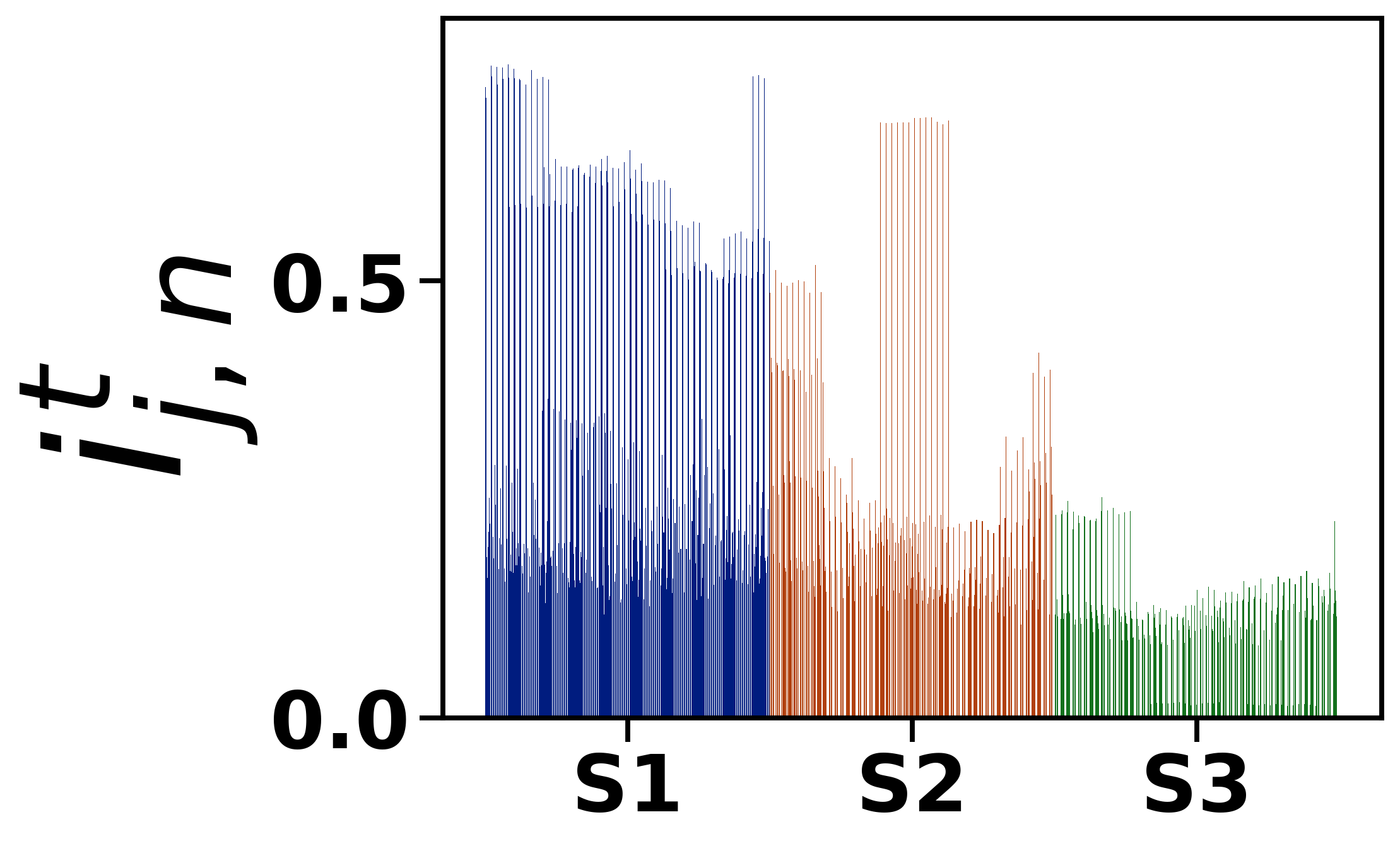}
& \includegraphics[width=0.33\columnwidth]{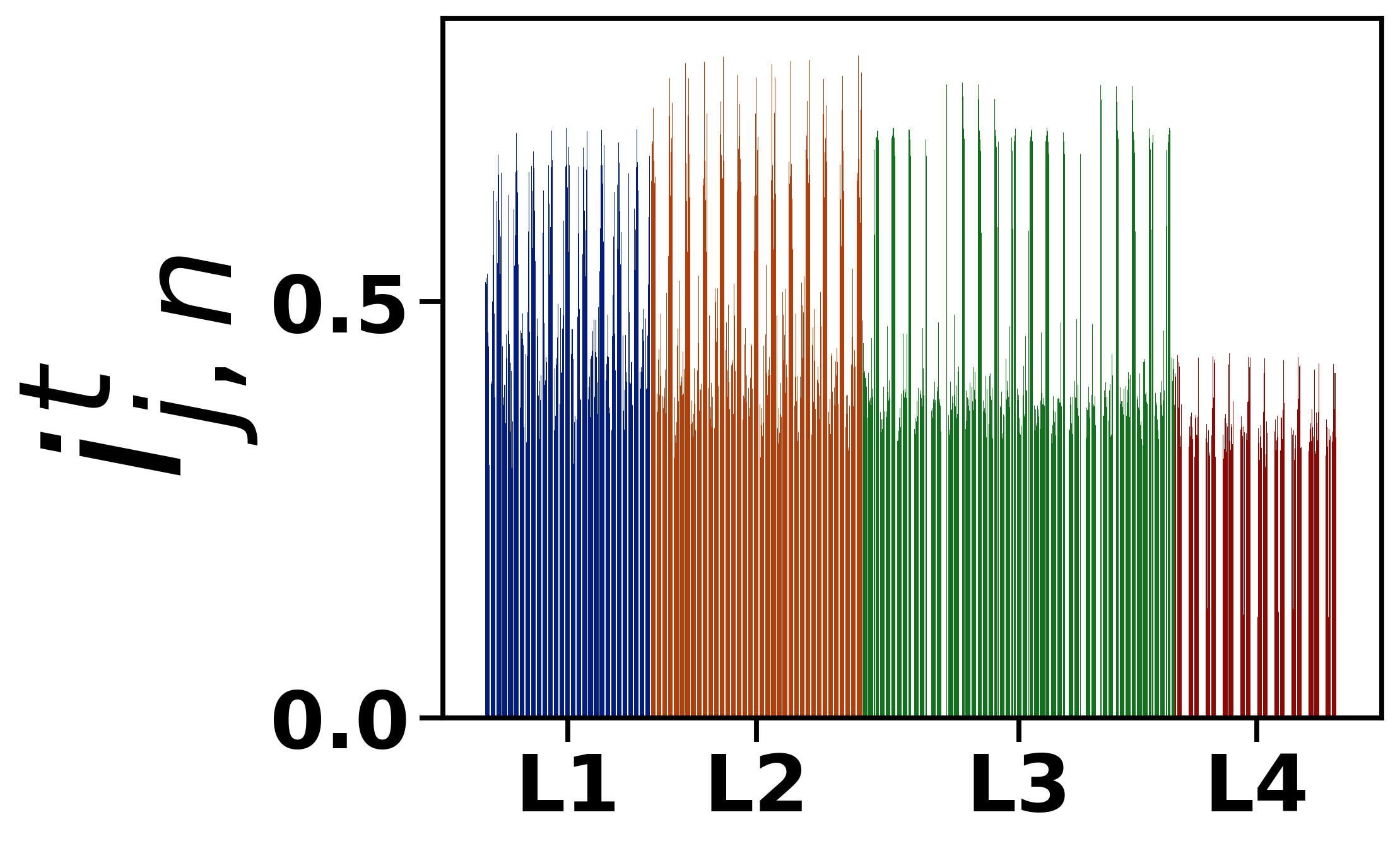} \\
\end{tabular}
\caption{Illustrations of LR importance $i_{j,n}^t$ across different convolutional blocks(B)/stages(S)/layers(L) during CTTA. [Top row] - Variations for the domain ``\textbf{glass\_blur}'' across datasets. [Bottom row] - Variations for the domain ``\textbf{snow}''.}
\label{fig: lr_imp}
\end{figure}

\noindent {\bfseries Analysis of LR importance $i_{j,n}^t$:} The prime objective of our work is to adaptively alter the LRs of parameters of selected layers based on their sensitivity. In Figure \ref{fig: lr_imp}, we illustrate how the LR importance $i_{j,n}^t$ (Eq. \ref{eqn:lr}) of WideResNet-28, ResNeXt-29, and ResNet-50, averaged for a task, change during adaptation on their respective datasets. For brevity, we show two domains - glass\_blur and snow. We observe that the importance varies across domains and models. Across all the models, we notice that generally, the deeper layers have smaller associated $i_{j,n}^t$, which means that the initial layers are adapted more to the target domain. This result completely aligns with the findings of surgical fine-tuning \cite{lee2022surgical}. In addition to that, we also notice more sparsity in the deeper layers. This further confirms the necessity of freezing parameters automatically selected based on the target domain for adaptation.

\noindent {\bfseries Layer importance via pseudo-labels:} LAW leverages pseudo-labels to determine layer importance for adaptation. We evaluate the effectiveness of this design for our method by investigating the utility of pseudo-label-based layer importance. We compute the gradients from the log-likelihood as $\nabla_{\theta_{j,n}^t} \log (p(x_k))$ and estimate the sensitivity of the parameters of the chosen layers (Eq. \ref{eqn:final_sens}-Eq. \ref{eq:d_ind}), and alter the LRs (Eq. \ref{eqn:lr}). Table \ref{table:pseudo} reports the mean classification errors for all the datasets. We observe that such gradients are unreliable indicators of prediction uncertainty. The scoring function $\mathcal{Z}_{\theta_{j,n}^t} = \|{{\nabla_{\theta_{j,n}^t} \mathcal{L}({\theta}^t)}}\|_1$ computed using these gradients carries insufficient information. We achieve improvements of 2.4\%, 3.7\%, and 8.5\% over CIFAR-10C, CIFAR-100C, and ImageNet-C, respectively.

\begin{table}[!t]
\centering
\scriptsize
\begin{minipage}[t]{0.48\linewidth}
\centering
\begin{tabular}{
@{}l
@{\hspace{1pt}}c
@{\hspace{1pt}}c@{}}
\toprule
\, \, \textbf{Dataset} & \textbf{Pseudo-labels} & \textbf{Ours} \\
\midrule
CIFAR-10C  & 17.9 & \textbf{15.5} \\
CIFAR-100C & 33.8 & \textbf{30.1} \\
ImageNet-C & 68.6 & \textbf{60.1} \\
\bottomrule
\end{tabular}
\caption{Prediction uncertainty using pseudo-labels vs. ours.}
\label{table:pseudo}
\end{minipage}%
\hfill
\begin{minipage}[t]{0.48\linewidth}
\centering
\begin{tabular}{
@{}l
@{\hspace{1pt}}c
@{\hspace{1pt}}c
@{\hspace{1pt}}c
@{\hspace{1pt}}c
@{\hspace{1pt}}c
@{\hspace{1pt}}c
@{\hspace{1pt}}c@{}}
\toprule
\,\, \, \, \, \, \textbf{$\lambda$} & \textbf{0} & \textbf{0.01} & \textbf{0.05} & \textbf{0.1} & \textbf{0.5} & \textbf{1.0} \\
\midrule
CIFAR-10C  & 16.8 & \textbf{15.5} & 16.4 & 17.0 & 17.6 & 17.9 \\
CIFAR-100C & 30.4 & \textbf{30.1} & 30.5 & 30.5 & 30.6 & 30.7 \\
ImageNet-C & 63.9 & \textbf{60.1} & 60.3 & 60.4 & 60.4 & 60.4 \\
\bottomrule
\end{tabular}
\caption{Ablation results with varying $\lambda$.}
\label{table:lambda}
\end{minipage}
\end{table}

\begin{figure}[htb!]
\centering
\scriptsize
\begin{tabular}{
@{}c
@{\hspace{0.1pt}}c@{}}
\includegraphics[width=0.5\linewidth]{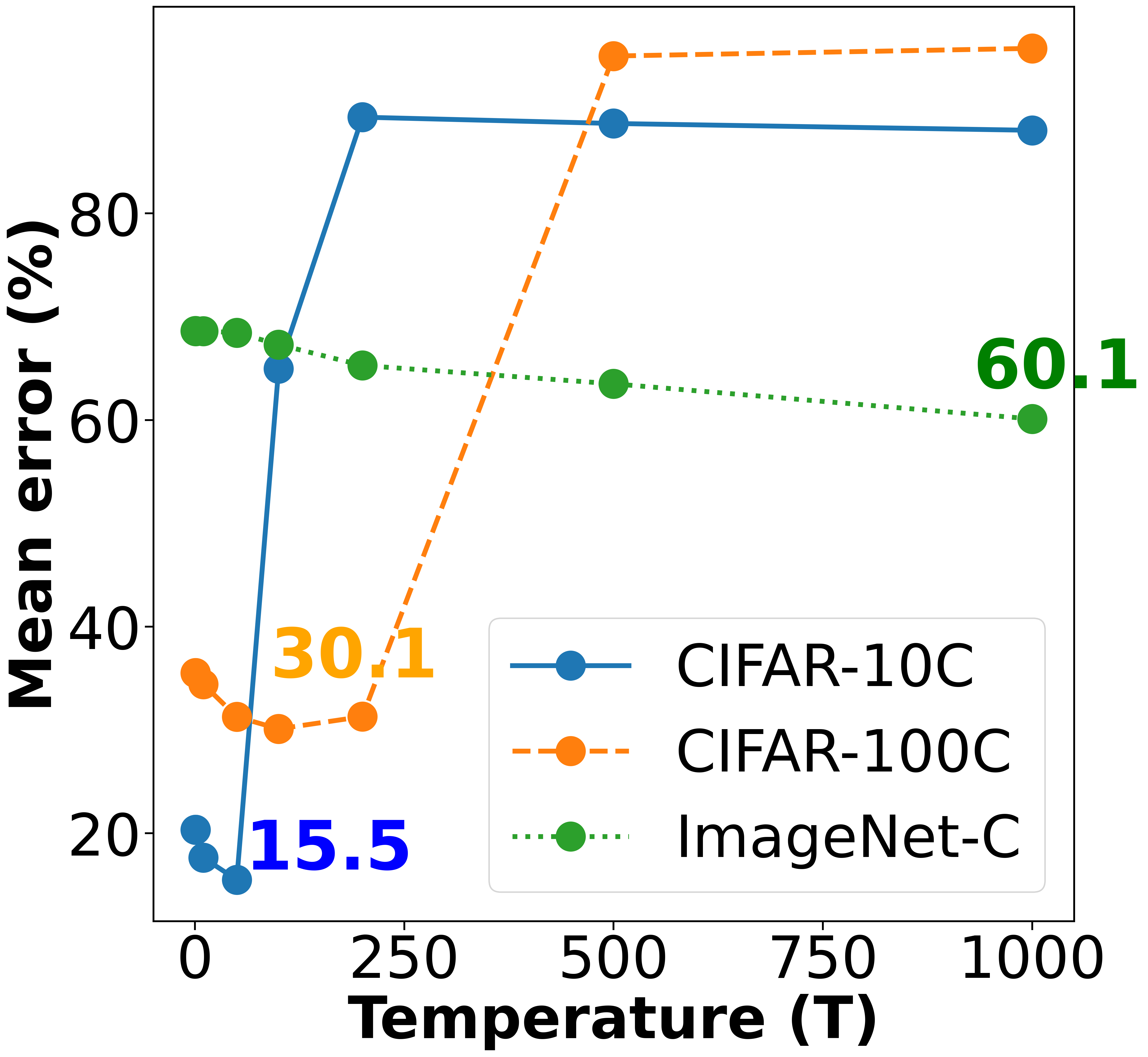} & 
\includegraphics[width=0.5\linewidth]{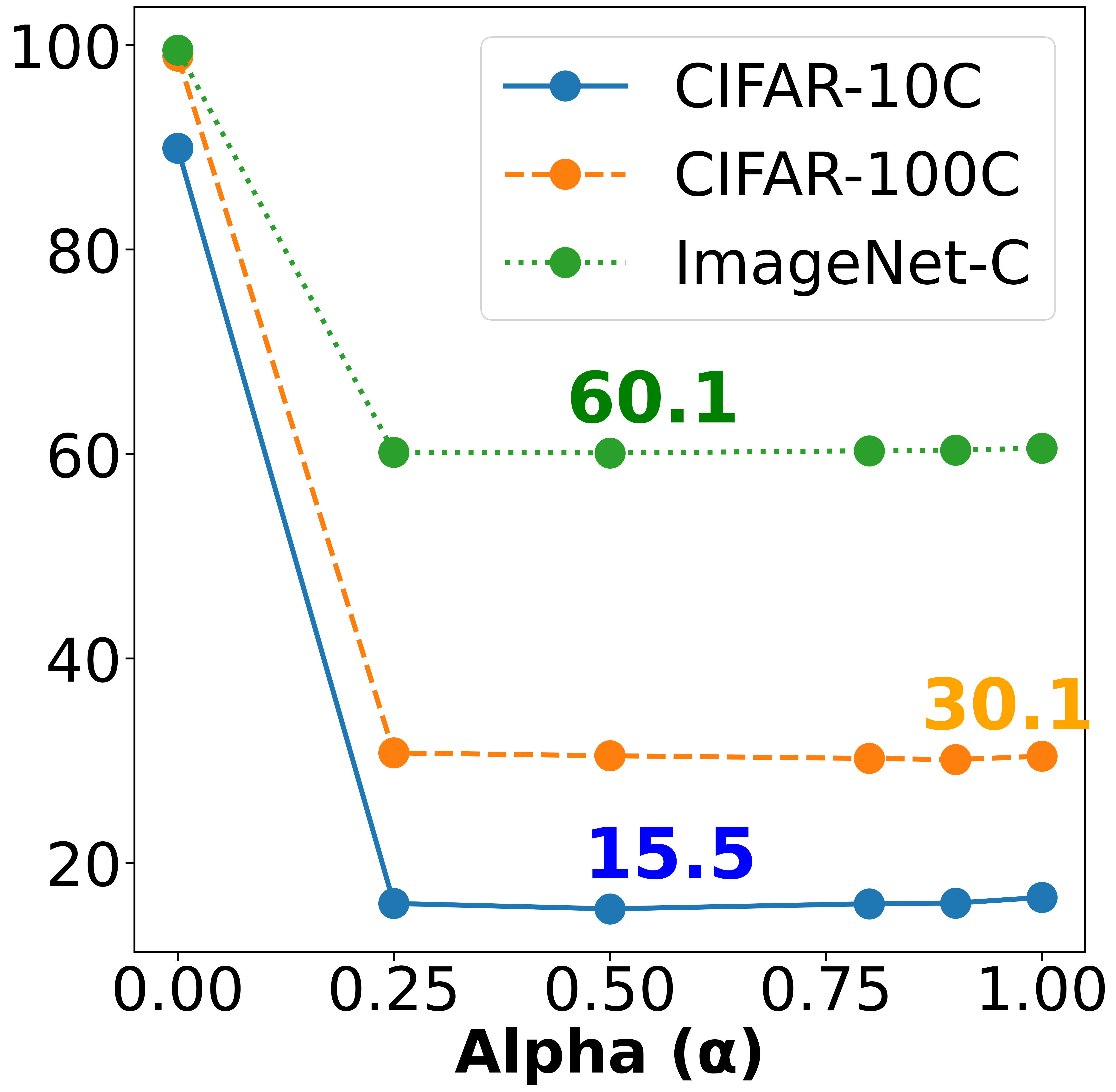} \\
\footnotesize{a) Ablation on temperature T} & 
\footnotesize{b) Ablation on $\alpha$} \\
\end{tabular}
\caption{Ablation results on smoothing factor $\alpha$ and temperature T.}
\label{fig:ablations}
\end{figure}

\noindent {\bfseries Effectiveness of temperature \textit{T}:} In our method, we compute the KL divergence between the uniform distribution \textbf{u} and a smoothed classifier's output logits (Eq. \ref{eq:initial}), denoted by $\hat h(x_k)$. We illustrate the results in Figure \ref{fig:ablations}(a) to study its influence. Ideally, larger values of \textit{T} push the distributions of the predictions closer to \textbf{u} \cite{hinton2015distilling} by spreading the probability among inputs. In CTTA, since we deal with domain shifts with no labels, we argue that higher values of \textit{T} are useful to capture the uncertainty. So, the KL divergence reduces, indicating more uncertainty in the model and hence, more adaptation.  For a systematic temperature selection, we aim to adjust \textit{T} based on the number of adapted parameters. In particular, if the number of adapted parameters increases across batches, \textit{T} will be increased. For a difficult dataset like ImageNet-C, larger values of \textit{T} are useful with the error reducing as \textit{T} increases. For the CIFAR-C datasets, the errors begin to rise for \textit{T} $\ge$ 100 which could be due to large spreads of the distributions.

\noindent {\bfseries Contribution of smoothing factor $\alpha$:} A key component of PALM is computing the moving average of sensitivities across data batches, smoothed by $\alpha$ (Eq. \ref{eq:unc}). Figure \ref{fig:ablations}(b) illustrates the variations of the mean errors, for the benchmark datasets, for varying $\alpha$. We observe that the ``domain-level" sensitivity across past batches of data is indeed useful. For $\alpha$ = 0 \textit{i.e.}, when $S_{j,n}^t$ of a selected parameter at timestep \textit{t} is not considered, the respective datasets' error rate drastically spikes. For CIFAR-10C, we achieve the best performance when $\alpha$ = 0.5 with an improvement of 1.7\% over LAW. For CIFAR-100C and ImageNet-C, we see improvements of 1.6\% and 0.9\% over LAW, at $\alpha$ = 0.9 and 0.5 respectively. With higher values of $\alpha$, the results do not vary too much. Hence, attaining a balance between sensitivities from past domains ($\hat S_{j,n}^t$) and current sensitivity ($S_{j,n}^t$) is needed.

\noindent {\bfseries Robustness to $\lambda$:} In Eq. \ref{eq:final_optim}, we set $\lambda$ $\in$ \{0, 0.01, 0.05, 0.1, 0.5, 1.0\}. Table \ref{table:lambda} reports the average classification errors on the benchmark datasets. For $\lambda$ = 0.01, we achieve the best results. In addition, we observe that our method is very robust to $\lambda$ \textit{i.e.}, with subtle variations to differing $\lambda$.

\section{Conclusion}
\label{conc}

In this paper, we propose a novel approach called PALM, for continual test-time adaptation. Our approach automatically decides on the importance of a layer by leveraging information from the gradient space \textit{i.e.}, by backpropagating the KL divergence between a uniform distribution and the softmax outputs. This effectively eliminates the need for pseudo-labels. For the parameters of these layers, selected based on the norm of the gradients, the LRs are adjusted based on their sensitivity to domain shift. Extensive experiments are conducted on benchmark datasets to demonstrate the efficacy of our method against different baselines. The results reveal that PALM achieves superior performance and pushes the state-of-the-art on several standard baselines.

\section{Acknowledgments} We thank all the reviewers for their helpful comments and suggestions. This project
was supported by a grant from the University of Texas at Dallas, Richardson, TX, USA.

\bibliography{aaai25}

\end{document}